\documentclass[letterpaper]{article} 
\usepackage{aaai23}  
\usepackage{times}  
\usepackage{helvet}  
\usepackage{courier}  
\usepackage[hyphens]{url}  
\usepackage{graphicx} 
\usepackage{natbib}  
\usepackage{caption} 
\usepackage{algorithm}
\usepackage{algorithmic}

\usepackage{newfloat}
\usepackage{listings}
\DeclareCaptionStyle{ruled}{labelfont=normalfont,labelsep=colon,strut=off} 
\floatstyle{ruled}
\newfloat{listing}{tb}{lst}{}
\floatname{listing}{Listing}
\title{Considerations for End-User Development in the Caregiving Domain}
 \author{
     Laura Stegner\textsuperscript{\rm 1}, 
     David Porfirio\textsuperscript{\rm 2}, 
     Mark Roberts\textsuperscript{\rm 3}, 
     Laura M. Hiatt\textsuperscript{\rm 3}
 }
 
 \affiliations{
     \textsuperscript{\rm 1}Department of Computer Sciences; University of Wisconsin--Madison, Madison, WI, 53706\\
     \textsuperscript{\rm 2}NRC Postdoctoral Research Associate; Naval Research Laboratory, Code 5514; Washington, DC, 20375\\
     \textsuperscript{\rm 3}Navy Center for Applied Research in AI; Naval Research Laboratory, Code 5514; Washington, DC, 20375

    \textsuperscript{\rm 1}stegner@wisc.edu, \textsuperscript{\rm 2}david.porfirio.ctr@nrl.navy.mil, \textsuperscript{\rm 3}\{mark.roberts, laura.hiatt\}@nrl.navy.mil
 }

\begin{document}

\maketitle

\begin{abstract}

As service robots become more capable of autonomous behaviors, it becomes increasingly important to consider how people communicate with a robot {\em what} task it should perform and {\em how} to do the task. Accordingly, there has been a rise in attention to \textit{end-user development} (EUD) interfaces, which enable non-roboticist end users to specify tasks for autonomous robots to perform. However, state-of-the-art EUD interfaces are often constrained through simplified domains or restrictive end-user interaction. Motivated by prior qualitative design work that explores how to integrate a care robot in an assisted living community, we discuss the challenges of EUD in this complex domain. One set of challenges stems from different user-facing representations, \textit{e.g.}, certain tasks may lend themselves better to rule-based trigger-action representations, whereas other tasks may be easier to specify via sequences of actions.
The other stems from considering the needs of multiple stakeholders, \textit{e.g.}, caregivers and residents of the facility may all create tasks for the robot, but the robot may not be able to share information about all tasks with all residents due to privacy concerns.  
We present scenarios that illustrate these challenges and also discuss possible solutions.
\end{abstract}

\begin{figure*}[!t]
    \centering
    \includegraphics[width=\textwidth]{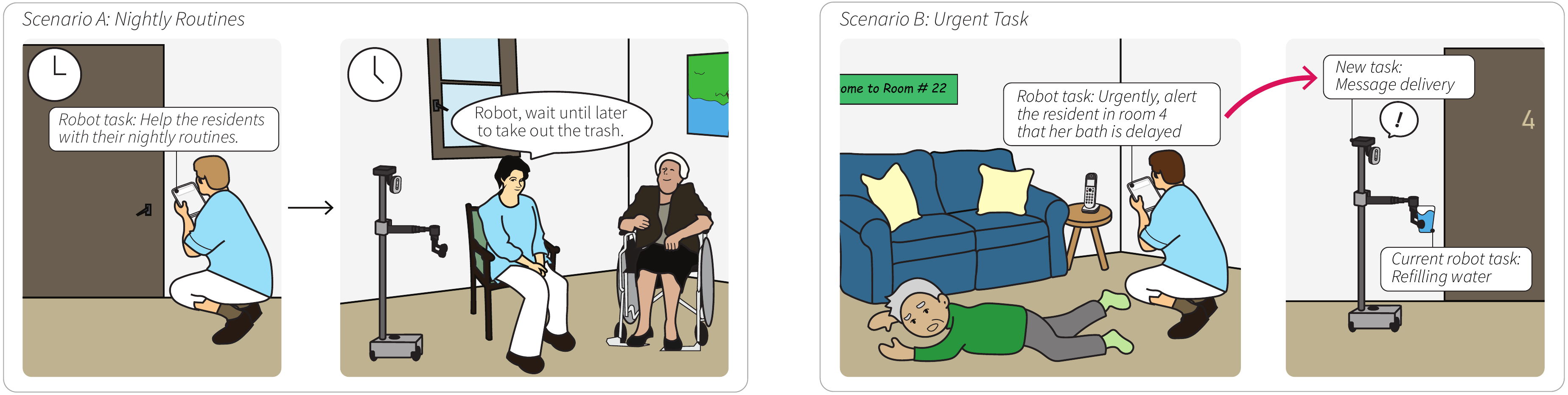}
    \caption{Scenarios from the caregiving domain, based on past work with caregivers and residents of an assisted living facility.}
    \label{fig:scenarios}
\end{figure*}

\noindent
\section{Introduction}
Robots are gradually becoming more ubiquitous in daily life. They can be found today performing roles such as food delivery directly to one's home \cite{jorgenson2023starship}, food service in restaurants \cite{chen2022non}, and customer service at airports \cite{ap2017meet}. As robots become more capable of autonomous behaviors, it is increasingly important to consider how non-roboticists can effectively communicate to robots the context-specific details of a task, including the desired task flow, behavioral parameters, and any task constraints. To address this need, the human-robot interaction (HRI) community has begun to focus on \textit{ end-user development} (EUD). EUD aims to allow end users, who are generally non-programmers, ``to adapt systems at a level of complexity that is appropriate to their individual skills and situations'' \cite{lieberman2006end}. 

Imagine the following scenario: an assisted living facility acquires a robot to help caregivers with their work. This robot has many capabilities, such as autonomous navigation, manipulation, and the ability to perceive and socially interact with its surroundings. Caregivers are able to assign tasks to the robot. Residents can also interact with the robot, often wanting to chat with it, make simple requests, or correct how the robot performs a task assigned by the caregiver.

In this example scenario, a highly capable autonomous robot must understand what tasks it should complete and the preferences of the stakeholders on how to do them. Pure autonomy (\textit{i.e.}, without stakeholders in the loop) will prove ineffective due to the uniqueness of caregiving settings, meaning that the robot will require copious context-based input from domain experts, developers, and other stakeholders in the field. Further, in domains such as caregiving, the tolerance for errors can be quite low due to factors such as safety concerns. Therefore, robot autonomy should be used in tandem with other techniques such as EUD that make use of the wealth of knowledge that end users possess. 

HRI research has explored EUD for a variety of domains such as manufacturing and at-home environments. This body of research includes many different robots such as mobile (\textit{e.g.}, \citet{huang2020vipo, porfirio2023sketching}), collaborative (\textit{e.g.}, \citet{paxton2017costar}), and social (\textit{e.g.}, \citet{pot2009choregraphe}), in addition to many different programming paradigms such as trigger action (\textit{e.g.}, \citet{ur2014practical}) and block based (\textit{e.g.}, \citet{huang2017code3}). For example, with the \textit{Vipo} EUD tool, users annotate a top-down floor map with the actions the robot must take \cite{huang2020vipo}. 

These EUD systems represent important steps towards democratizing the use of robots in everyday life. However, when robots are more widely used, they will be faced with more complex and unconstrained environments than the currently researched use cases encompass. 

Caregiving is one examplar domain that presents many interesting considerations for EUD. Caregiving is especially interesting due to the large discrepancies between caregivers and residents in terms of needs and capabilities. For example, residents may have cognitive or physical challenges and require slower-paced interactions. Caregivers, in contrast, are often short on time and seek quick and efficient interactions. In addition, caregiving presents numerous safety and privacy considerations, such as HIPAA laws that restrict the sharing of personal health information.

The EUD needs of the caregiving domain are not addressed by current systems. This position is supported by the observations and experience of the first author in conducting qualitative design studies in a local assisted living facility in the Midwestern United States. This previous work invovled several months spent working with both the older adult \textit{residents} \cite{stegner2023situated} and professional \textit{caregivers} \cite{stegner2022designing} to explore how robots can be integrated into the existing resident lifestyles and caregiver workflows. These design insights motivate discussion on EUD for the caregiving domain.

In what follows, we present background information on two key challenges related to EUD in the caregiving domain. We situate these challenges within two fictional caregiving scenarios. For each scenario, we identify the challenges and discuss possible solutions. We conclude by discussing future work on EUD and caregiving.


\section{Background}
We provide background on two key EUD challenges facing the caregiving domain: (1) \textit{ user-facing representations} and (2) \textit{stakeholders of the caregiving domain}.

\subsection{User-Facing Representations}
A \textit{user-facing representation} refers specifically to the way that a robot's task logic is presented to users for viewing and interacting. Current EUD research has explored a variety of different ways of representing this logic. Within the caregiving domain, different stakeholders may benefit from different representations depending on what kind of task they are attempting to specify or personalize for the robot to perform. From \citet{stegner2022designing} and \citet{stegner2023situated}, we observe that these tasks could naturally fit into the following representational forms.

\textit{Trigger-Action Programs (TAP)}.\ With TAP, end users specify individual rules of the form ``IF [trigger] AND [condition] THEN [action]'' \cite{ur2014practical}. TAP rules are easily expressed in natural language, as in the following examples:
\begin{itemize}
    \item ``If the trash is full, empty it''
    \item ``If the resident asks for more water, give it to them''
    \item ``If there's a dirty dish on the table, and there is nobody sitting at that table, take the dish to the kitchen''
\end{itemize}
TAP rules can be simple, and users provide any number of them to express a robot's responses to disjoint triggers. However, TAP rules are interpreted as individual and isolated program constraints, so TAP end-user developers cannot express linear sequences of events and more complex logic. 

\textit{Scheduled/recurring activities}.\ Recurring activities are a special form of TAP in which the trigger is always based on time. For example:
\begin{itemize}
    \item ``Every Monday at 3pm, escort Jim to the group activity.''
    \item ``Make coffee every day at 6am and 2pm.''
    \item ``At 8am daily, if I am not awake, wake me up.
\end{itemize}

\textit{Plan}.\ A plan consists of a sequence of actions for the robot to complete, \textit{e.g.}, for a morning routine: (1) start the coffee maker; (2) make toast; (3) bring me coffee and toast; (4) when I am done, clean up. Plans in the caregiving domain may also include branching or looping logic.


\subsection{Stakeholders of the Caregiving Domain}
The caregiving domain has many stakeholders, such as professional caregivers and nurses who provide daily care to older adult residents \cite{stegner2022designing}. The caregiver may give the robot care tasks such as checking on residents, delivering food or medication, or clearing dirty dishes, whereas the residents themselves may want to personalize the way the robot performs tasks for them or ask it to perform tasks that are outside of a caregiver's duties (\textit{e.g.}, picking up a library book or rearranging items on a shelf). Additional stakeholders who can specify or personalize robot tasks include family or friends of residents and other facility staff, such as administrators and activity coordinators \cite{calvaresi2017exploring}.

Each complex domain may have its own challenges specific to the different stakeholders involved. Through past work, we have developed an understanding of two key stakeholders in the caregiving domain: caregivers \cite{stegner2022designing} and residents \cite{stegner2023situated}. Although many other stakeholders exist within the caregiving domain, starting with just caregivers and residents provides information on the challenges that may arise from accommodating additional stakeholders.

\section{Scenarios}

The following fictional scenarios are inspired by our previous observations at the assisted living facility. Each scenario is depicted in Figure \ref{fig:scenarios}.

\subsection{Scenario A}
Imagine Caregiver A, who tries to create a task for the robot to assist multiple residents with their bedtime routines, including trash pickup. Caregiver A needs to specify several levels of information: prioritization of which residents must be helped before others, the goals or actions associated with helping each resident, and other task parameters such as how social the robot should be. Then, while the robot is in one of the resident's rooms following Caregiver A's task specification, it notices another caregiver, Caregiver B, who is already in the room with the resident. Caregiver B is helping the resident sort through her mail. Since most of the mail is spam, it will go to the trash. Caregiver B asks the robot to wait until later to empty the trash, since Caregiver B is currently using it to sort the mail.

The scenario above presents challenges related to user-facing representations and receiving a multitude of possibly conflicting task input from each stakeholder. Caregiver A must specify the robot's task including many different levels of detail from the high-level flow of which residents to visit and when, down to the fine-grained details of individual interactions with each resident. Caregiver B has the challenge of modifying the robot's existing task during its execution.

\subsection{Scenario B}
In our previous observations and experience in the field, we found that residents may become agitated if their routines are disrupted. Residents are also often curious about the affairs of the facility, particularly when another resident is sick or injured. On the basis of those observations, we present the following scenario.

Imagine that one human caregiver, Caregiver A, creates a task for the robot to refill water for all of the residents every afternoon. One afternoon, the robot is performing this scheduled task. Elsewhere in the facility, one resident, Resident A, is scheduled for Caregiver A to assist them with a shower. However, Caregiver A suddenly receives an alert that another resident, Resident B, fell and needs immediate assistance. Resident A is very anxious and will become agitated if Caregiver A is late, so Caregiver A creates an urgent task for the robot to inform Resident A that the shower is delayed due to an emergency. Resident A is curious about the emergency that caused the delay, wanting to know who had an accident. Although the robot may have knowledge of which resident fell, it cannot disclose this information to the curious resident.

The scenario above presents two challenges related to the involvement of multiple stakeholders. 
The first challenge relates to privacy---Resident A asks the robot for information that the robot should not share to protect Resident B's privacy. The scenario also presents a challenge about priority conflicts between tasks---the robot is currently executing a task from one caregiver when it is given another urgent task to complete by another.

\section{Considerations for EUD for Caregiving}
We use the two scenarios presented above to motivate four considerations for EUD for caregiving.

\subsection{Supporting Specification at Different Levels of Detail}
The caregiver in Scenario A must specify many levels of detail in order to configure the robot to help the residents get ready for bed. Interfaces currently may account for one of these levels of detail, whereas a caregiver needs to be able to access all of them. We see in Scenario A that the caregiver must specify, for example, the schedule of which residents to visit and when, as well as the task-level details of what the robot should do for that resident. The caregiver may also need to specify context-specific details about how the robot should interact, such as whether the robot should chat with the resident or if it should avoid making noise because the resident is not feeling well.

One possible solution to this challenge is a hierarchical interface that allows caregivers access to different levels of detail when appropriate. Caregivers need to be able to easily switch between different levels of detail so that they can have full control over the robot's behavior. From past work by  \citet{stegner2022designing}, we find that caregivers want to ensure that the robot is doing the right task in the correct way so that the residents are safe and happy. Since caregivers are intimately knowledgeable about residents' routines, needs, and tendencies, they need to be able to express all of this domain knowledge to the robot. For example, a caregiver would know the best order to help residents get ready for bed. A caregiver may also know for each resident how they want a task to be executed, such as indicating how conversational a particular resident is feeling on a specific day. While the robot could learn personalized behaviors over time, \citet{stegner2023situated} observed that the residents grow frustrated when they have to constantly repeat how something should be done.

\subsection{Modifying Existing Tasks during Execution}
In Scenario A, the robot begins to complete the task specified by Caregiver A. However, upon arrival, the robot finds Caregiver B is already present. Because Caregiver B is already helping the resident prepare for bed, Caregiver B needs to modify the robot's task based on what remains to be done. Since the robot is already executing the task, this situation requires that Caregiver B's input be inserted into the execution of the task in real time. 

For the task to be modified in real time, the representations presented to both caregivers must be compatible. For example, if Caregiver A specifies a plan for what the robot should do in that resident's room, Caregiver B may add, remove, or update steps within that plan. If Caregiver A instead specifies the task in terms of TAP rules, then the TAP representation may need to be extended to enable modification by Caregiver B. Skipping or canceling triggers is trivial, but adjusting the robot's response in real time or deferring its actions until a later time may not be simple with a TAP representation. 
Therefore, it is important to consider how we can convert between different user-facing representations to allow users to view and interact with the most appropriate one for their goals.


\subsection{Access to Information}
From Scenario B, when the robot tells the resident that their shower is delayed due to an emergency, the resident wants to know more details. The robot must understand what information it can share with the curious resident asking about their neighbor. Because multiple users may be creating tasks for or merely interacting with the robot, there will be instances when it is not appropriate to share such information with every user. In fact, HIPAA laws in the United States restrict the sharing of personal health information. From our scenario above, even if the robot possesses information about the other resident's fall, it is not allowed to share it. Otherwise, it violates the resident's privacy. 

Information within the EUD system needs to be clearly marked according to who can access it. However, asking for all information to be labeled individually would be too cumbersome for caregivers and complicated for residents. Instead, different classes of stakeholders could have different permissions when interacting with the robot. For example, a resident might be able to see an anonymized version of the tasks the robot has scheduled, but can only access or edit information about tasks that directly involve them. Caregivers, on the other hand, need access to all details for all of the robot's tasks, and may even need the ability to block edits from certain residents when their health or safety could be compromised (\textit{e.g.}, the resident should not be permitted to cancel their medication delivery).

\subsection{Priority Conflicts}
Scenario B also presents a priority conflict due to the different tasks created by multiple caregivers. The robot has the scheduled task to refill water that it is currently executing and the incoming task to inform the resident of the delay in their shower. The robot could queue the new task to be performed after all the water had been refilled. However, due to the urgency of the task, it is likely preferred for the robot to interrupt the water refilling task to pass on the message. 

Depending on the robot's task load, the robot may not even be able to feasibly complete all of the tasks assigned to it. The most naive approach to handling challenges relating to conflicting tasks is to use a first-come first-serve approach. However, this solution will easily break down if, such as in our example, the robot is busy with scheduled tasks, but the caregiver urgently needs the robot to complete another task instead.

Perhaps a better solution is to look at using automated scheduling techniques, such as oversubscribed scheduling. This technique is used when there are not enough resources available to complete every task, so certain tasks must only be partially completed or dropped. However, it may be necessary to ensure certain critical tasks always get done or some urgent task becomes the highest priority, so the task representations may require slight modifications to provide additional information to the scheduler about urgency and priority. This solution also indicates the need for another interface to manage the overarching schedule and allow end users to handle scheduling conflicts that could not be automatically resolved.

\section{Discussion}
Creating EUD solutions for the caregiving domain requires considerations beyond the constrained EUD tools that exist today. This paper presents scenarios that are inspired by previous qualitative research aimed at understanding how to integrate robotic assistance into assisted living facilities. These scenarios highlight some unique challenges of end-user development (EUD) for caregiving scenarios.

We were able to identify the challenges for EUD in the caregiving environment only due to this previous qualitative research. Through these studies, we became highly knowledgeable about the stakeholders and environment. As we think about EUD tools for more complex domains, it will be increasingly necessary to first understand the target domain. What use cases will the robot face? What stakeholders would need to give it tasks? What are those stakeholders' needs, abilities, and preferences?

Human-Computer Interaction (HCI) researchers have a rich history of working with stakeholders through participatory design methods to understand how to design technology for specific users (\textit{e.g.}, \citet{gronvall2013participatory, scandurra2013participatory}). These methods have more recently become popular among HRI researchers, although the primary focus is on designing robotic interventions or interactions (\textit{e.g.}, \citet{lee2017steps, broadbent2009retirement}). Systems researchers looking to build EUD tools for specific domains could benefit from similar participatory methods.

Once EUD tools are developed, it is also a challenge to sufficiently test them. Laboratory testing is limited, but testing in the wild can be tricky. In our case, it is not necessarily feasible to bring older adults (due to mobility or safety concerns, for example) or caregivers (due to limited time) to a controlled laboratory space to test out interfaces. Instead, we will only be able to solicit stakeholder feedback by taking the interfaces to them.

In-the-wild evaluations pose many challenges of their own, particularly when trying to evaluate a prototype system that are not robust to less controlled environments. However, it is still important to put systems in front of end users. These users are the true experts in the target domain, so they can to provide rich feedback on its usability and usefulness.

We suggest overall that EUD is a necessary tool to integrate autonomous robots into complex domains. To fully understand the needs of interfaces for these complex domains, it is critical to take time to understand the end users, their needs, and the complexities of the environment.

\section{Acknowledgements}
We would like to thank the residents and caregivers who participated in our research. 
This material is based upon work supported by National Science Foundation (NSF) award IIS-1925043 and the NSF Graduate Research Fellowship Program under Grant No. DGE-1747503. 
This research was also supported by the Office of Naval Research. The views and conclusions contained in this document are those of the authors and should not be interpreted as necessarily representing the official policies, either expressed or implied, of the US Navy or the NSF.

\bibliography{ref}

\end{document}